\documentclass[]{interact}

\usepackage{epstopdf}

\usepackage[doublespacing]{setspace}

\usepackage{natbib}

\usepackage{appendix}
\usepackage{fancyhdr}
\usepackage{float}

\usepackage{booktabs}
\usepackage{multirow}
\usepackage{algorithm}
\usepackage{algpseudocode}

\usepackage{mathtools}
\usepackage{amsmath, amssymb, bbm}

\usepackage{graphicx}
\usepackage{subcaption}
\usepackage{tikz}
\usetikzlibrary{positioning}
\usepackage{hyperref} 

\begin{document}

\articletype{ARTICLE TEMPLATE}

\title{Counterfactual Survival Q-learning via Buckley-James Boosting, with Applications to ACTG 175 and CALGB 8923}

\author{
\name{Jeongjin Lee\textsuperscript{a} and Jong-Min Kim\textsuperscript{b}\thanks{Corresponding Author: Jong-Min Kim. Email: jongmink@morris.umn.edu}}
\affil{\textsuperscript{a}Division of Biostatistics, The Ohio State University, 281 W Lane Ave, Columbus, OH 43210, U.S.A.; 
\textsuperscript{b}Division of Science and Mathematics, University of Minnesota-Morris, Morris, MN 56267, U.S.A.; EGADE
Business School, Tecnol\'ogico de Monterrey, Ave. Rufino Tamayo, Monterrey 66269, Mexico.}
}

\maketitle

\begin{abstract}
We propose a Buckley–James (BJ) Boost Q-learning framework for estimating optimal dynamic treatment regimes from right censored survival outcomes in longitudinal randomized clinical trials, motivated by the clinical need to support patient specific treatment decisions when follow up is incomplete and covariate effects may be nonlinear. The method combines accelerated failure time modeling with iterative boosting using flexible base learners, including componentwise least squares and regression trees, within a counterfactual Q-learning framework. By modeling conditional survival time directly, BJ Boost Q-learning avoids the proportional hazards assumption, yields clinically interpretable time scale contrasts, and enables estimation of stage specific Q-functions and individualized decision rules under standard potential outcomes assumptions. In contrast to Cox based Q-learning, which relies on hazard modeling and can be sensitive to nonproportional hazards and model misspecification, our approach provides a robust and flexible alternative for regime learning. Simulation studies and analyses of the ACTG175 HIV trial and the CALGB 8923 two stage leukemia trial show that BJ Boost Q-learning improves treatment decision accuracy and produces more stable within participant counterfactual contrasts, particularly in multistage settings where estimation error and bias can compound across stages.
\end{abstract}


\section{Introduction}
\label{s:intro}

\sloppy

In the evolving field of contemporary healthcare, individualized treatment strategies \citep{moodie2012q, wahed2013evaluating, chakraborty2014dynamic, kosorok2015adaptive, song2015penalized, simoneau2020estimating, cho2023multi} have become a central approach for tailoring interventions to optimize patient outcomes. This approach is especially important in the presence of complex censored data, where event times may be partially unobserved due to issues such as patient dropout or loss to follow-up. Such challenges have motivated the development of methodological frameworks capable of handling incomplete outcome information.

Survival analysis, which focuses on time-to-event data, is crucial in medical research. A major challenge in this field is the frequent occurrence of censoring, where the event time for certain observations is unknown, making it difficult to make well-informed decisions from these datasets. Cox regression \citep{cox1972regression} is commonly used for analyzing time-to-event data with censoring. However, it has limitations, including the assumption of proportional hazards and a less intuitive interpretation compared to linear regression. Moreover, hazard ratios (HRs) derived from the Cox model are relative measures that compare the hazard rates of two groups and depend on the underlying hazard function, which may not always be straightforward to interpret. The Cox PH model assumes that the hazard function for an individual is a product of a baseline hazard function and an exponential function of a linear combination of the covariates. This linearity assumption may not hold in all situations, especially when the relationships between covariates and the hazard are complex and non-linear.

An alternative approach is the accelerated failure time (AFT) model \citep{bu79, wei1992accelerated, ji03, ji06, ch21}, which offers several advantages. The AFT model directly models survival time using a linear regression form, providing more interpretable effects than hazard ratios in Cox models. Moreover, the AFT model is robust to violations of the proportional hazards assumption. The Buckley–James (BJ) method \citep{bu79, ji06}, a semiparametric estimator for the AFT model, accommodates arbitrary censoring mechanisms and remains consistent under mild conditions. Due to these advantages, several penalized extensions of BJ estimation have been developed to address high-dimensional data \citep{jo08, wa08, jo09, li14, lee2024censored}.

To extend the Buckley–James (BJ) framework to non-linear and high-dimensional settings, \citet{wang2010buckley} proposed the BJ Boosting algorithm, which iteratively updates the predictive function using flexible base learners such as componentwise least squares or regression trees. 
Unlike Cox-based machine learning approaches that model hazard functions, BJ Boosting directly targets the log-transformed survival time under the accelerated failure time (AFT) model, enabling more interpretable and robust inference. 
It effectively accommodates right-censoring and tied event times while capturing complex, non-linear relationships. In terms of computational efficiency and estimation accuracy, BJ Boosting has shown superior performance over classical BJ and Cox models. 
Its iterative structure adaptively selects relevant covariates and mitigates overfitting, making it particularly well-suited for biomarker-driven survival analysis and dynamic treatment regime estimation.

Reinforcement learning, particularly Q-learning \citep{wa92}, has shown great potential in personalized treatment optimization due to its ability to adapt and learn from sequential decision-making processes. In healthcare, treatment decisions are often made in stages, considering the evolving state of a patient's health. Q-learning can model this dynamic process by learning optimal policies that maximize long-term health outcomes based on cumulative rewards. This adaptive capability makes Q-learning especially promising for personalized treatment, where the goal is to tailor interventions to individual patient needs over time. Despite its promise in personalized treatment optimization, directly applying Q-learning to censored survival data proves challenging due to severe censoring.

To address these challenges, we propose the BJ Boost Q-learning under a counterfactual framework, which integrates Q-learning with the Buckley–James (BJ) boosting algorithm to accommodate nonlinear relationships under right censored survival data.
This approach is motivated by real world clinical studies spanning both single stage and multistage treatment settings.
As a single stage example, we consider the ACTG 175 trial, a randomized clinical trial designed to compare monotherapy with zidovudine or didanosine against combination therapies involving zidovudine and didanosine or zidovudine and zalcitabine, in HIV positive adults with CD4 T cell counts between 200 and 500 cells/mm\textsuperscript{3} \citep{hammer1996trial}.
As a multistage example, we consider the CALGB 8923 trial, a two stage randomized study in acute myeloid leukemia, where participants were randomized at baseline and a subset were rerandomized after achieving complete remission \citep{stone1995granulocyte, stone2001postremission}.
In such trials, each patient is assigned to one treatment arm at each decision point, but evaluating optimal dynamic treatment regimes requires estimation of potential outcomes under all possible treatment strategies, which necessitates counterfactual reasoning.
While \citet{lee2025qlearninglinear} introduced a linear BJ Q-learning approach under a counterfactual framework, its reliance on linear associations between covariates and survival outcomes limits its applicability in complex real world clinical contexts.
Furthermore, the linear Buckley–James estimator \citep{bu79, ji06} may exhibit convergence issues when the true data generating mechanism is nonlinear.
By employing flexible base learners such as componentwise least squares or regression trees, the BJ boosting approach enables stable and accurate imputation of censored survival times even in nonlinear settings.
Coupled with the recursive structure of Q-learning, this iterative boosting mechanism adaptively refines treatment value estimation at each stage, thereby improving the reliability of learned dynamic treatment regimes and ultimately enhancing patient specific clinical decision making.

\section{Method}

\subsection{Setup and notation}

We consider a sequentially randomized clinical trial with \(K\) decision stages indexed by \(k=1,\ldots,K\).
Stage \(k\) corresponds to the time interval from the stage \(k\) decision time to the earlier of the next decision time or a failure event.
For participant \(i\), let \(B_{i,0}\) denote baseline covariates, and let \(X_{i,k}\) denote time varying covariates measured immediately before the treatment decision at stage \(k\).
Let \(A_{i,k}\in\mathcal A\) denote the randomized treatment assigned at stage \(k\).
Define the pretreatment history at stage \(k\) as
\begin{equation}
H^{-}_{i,k}
=
\bigl(B_{i,0},X_{i,1},A_{i,1},\ldots,X_{i,k}\bigr),
\end{equation}
and define the full history under a candidate treatment \(a_k\in\mathcal A\) as
\begin{equation}
H_{i,k}(a_k)
=
\bigl(H^{-}_{i,k},a_k\bigr).
\end{equation}

Let \(T_{i,k}\) denote the true stage specific duration at stage \(k\), measured from the stage \(k\) decision time to the stage \(k\) endpoint, defined as the earlier of the next decision time and the failure event time.
The next decision time is the protocol specified decision point for stage \(k+1\), which may be visit based or event based and is observed only for participants who reach it without prior failure or censoring.
Let \(D_{i,k}\in\{0,1\}\) indicate the type of the stage \(k\) endpoint, where \(D_{i,k}=1\) if the failure event occurs at or before the next decision time and \(D_{i,k}=0\) if the next decision time occurs first.
Let \(C_i\) denote the overall censoring time measured from baseline.
To connect \(C_i\) to stage specific outcomes, define the elapsed time before stage \(k\) as
\begin{equation}
S_{i,1}=0,
\qquad
S_{i,k}=\sum_{\ell=1}^{k-1}T_{i,\ell},
\quad k\ge 2,
\end{equation}
and define the remaining censoring time at the beginning of stage \(k\) as
\begin{equation}
C_{i,k}
=
\max\{C_i-S_{i,k},\,0\}.
\end{equation}
Conditional on entering stage \(k\), the observed stage specific time and censoring indicator are
\begin{equation}
Y_{i,k}
=
\min(T_{i,k},C_{i,k}),
\qquad
\delta_{i,k}
=
\mathbb I(T_{i,k}\le C_{i,k}).
\end{equation}
Thus, \(\delta_{i,k}=1\) indicates that the stage endpoint is observed, in which case the endpoint type \(D_{i,k}\) is also observed, while \(\delta_{i,k}=0\) indicates that censoring occurs before the stage endpoint.
We observe stage \(k\) outcomes only when participant \(i\) has not failed and has not been censored before entering stage \(k\).
Let
\begin{equation}
\eta_{i,1}
=
\mathbb I(C_{i,1}>0),
\qquad
\eta_{i,k}
=
\mathbb I\bigl(\eta_{i,k-1}=1,\ \delta_{i,k-1}=1,\ D_{i,k-1}=0\bigr),
\quad k\ge 2,
\end{equation}
denote the indicator that participant \(i\) enters stage \(k\).

Let \(\bar a_k=(a_1,\ldots,a_k)\) denote a treatment history through stage \(k\), and let \(\bar a_0\) denote the empty history.
For each \(k\), let \(T_{i,k}(\bar a_k)\) denote the potential stage specific time at stage \(k\) under history \(\bar a_k\), and let \(D_{i,k}(\bar a_k)\) denote the corresponding potential stage endpoint type.
We treat the overall censoring time \(C_i\) as defined on the same baseline time scale as the potential stage times, so that counterfactual remaining censoring times can be obtained by subtracting counterfactual elapsed times from \(C_i\).
Define the potential elapsed time before stage \(k\) under history \(\bar a_{k-1}\) as
\begin{equation}
S_{i,1}(\bar a_0)=0,
\qquad
S_{i,k}(\bar a_{k-1})
=
\sum_{\ell=1}^{k-1}T_{i,\ell}(\bar a_\ell),
\quad k\ge 2,
\end{equation}
and define the potential remaining censoring time at the beginning of stage \(k\) as
\begin{equation}
C_{i,k}(\bar a_{k-1})
=
\max\{C_i-S_{i,k}(\bar a_{k-1}),\,0\}.
\end{equation}
Let \(\eta_{i,k}(\bar a_{k-1})\) denote the indicator that participant \(i\) reaches stage \(k\) under history \(\bar a_{k-1}\), defined by
\begin{align}
\eta_{i,1}(\bar a_0)
&=
\mathbb I\bigl\{C_{i,1}(\bar a_0)>0\bigr\}, \nonumber\\
\eta_{i,k}(\bar a_{k-1})
&=
\mathbb I\Bigl\{
\eta_{i,k-1}(\bar a_{k-2})=1,\ 
T_{i,k-1}(\bar a_{k-1})\le C_{i,k-1}(\bar a_{k-2}),\nonumber\\
&\hspace{7.2em}
D_{i,k-1}(\bar a_{k-1})=0
\Bigr\},
\qquad k\ge 2.
\label{eq:eta_reach_stage}
\end{align}
We define the cumulative survival time under a full treatment history \(\bar a=\bar a_K\) as
\begin{equation}
\label{eq:cum_surv}
T_{i,\mathrm{cum}}(\bar a)
=
\sum_{k=1}^{K}\eta_{i,k}(\bar a_{k-1})\,T_{i,k}(\bar a_k),
\end{equation}
so that only stages reached under \(\bar a\) contribute to the total.

We define optimal stage specific decision rules via backward recursion using Q-functions.
For \(k=1,\ldots,K\), define the true Q-function
\begin{align}
\label{eq:true_Q}
Q_k^{*}\!\bigl(H_{i,k}(a_k)\bigr)
&=
\mathbb E\Bigl[
T_{i,k}(\bar A_{i,k-1},a_k)
+
\mathbb I\bigl\{\eta_{i,k+1}(\bar A_{i,k-1},a_k)=1\bigr\}\nonumber\\
&\hspace{2.8em}\times
\max_{a_{k+1}\in\mathcal A}
Q_{k+1}^{*}\!\bigl(H_{i,k+1}(a_{k+1})\bigr)
\ \Bigm|\ 
H_{i,k}(a_k),\ 
\eta_{i,k}(\bar A_{i,k-1})=1
\Bigr].
\end{align}
with \(Q_{K+1}^{*}\equiv 0\), where \(\bar A_{i,k-1}=(A_{i,1},\ldots,A_{i,k-1})\) denotes the realized treatment history through stage \(k-1\).
The indicator \(\mathbb I\{\eta_{i,k+1}(\bar A_{i,k-1},a_k)=1\}\) ensures that continuation value contributes only when stage \(k+1\) is reached under \((\bar A_{i,k-1},a_k)\).
The optimal decision at stage \(k\) is
\begin{equation}
d_{k}^{*}(H^{-}_{i,k})
=
\arg\max_{a_k\in\mathcal A} Q_k^{*}\!\bigl(H_{i,k}(a_k)\bigr).
\end{equation}

\subsection{Buckley–James boost Q-learning}

For stage specific right censored outcomes, we use Buckley--James boosting \citep{wang2010buckley, wang2023package} to impute censored stage times and to construct the pseudo outcomes used for backward Q-learning.
Within each stage \(k\), we assume stage specific conditional independent censoring given the stage history.
Using potential outcome notation, this condition can be expressed as
\begin{equation}
\label{eq:stage_censor_indep}
T_{i,k}(\bar a_k)\ \perp\ C_{i,k}(\bar a_{k-1})
\ \big|\ H_{i,k}(\bar a_k),\ S_{i,k}(\bar a_{k-1}),\ \eta_{i,k}(\bar a_{k-1})=1,
\qquad \text{for all }\bar a_k\in\mathcal A^k.
\end{equation}
Here \(H_{i,k}(\bar a_k)\) denotes the covariate and treatment history that would be observed through the stage \(k\) decision time under history \(\bar a_k\), and \(S_{i,k}(\bar a_{k-1})\) is the elapsed time from baseline to the stage \(k\) decision time under \(\bar a_{k-1}\).

The imputation uses a regression function \(f_{k,a}(\cdot)\) for the conditional mean of a working response.
For clarity, we describe the procedure for a generic response \(Y_{i,k}\).
In applications we may take \(Y_{i,k}=\log\{\min(T_{i,k},C_{i,k})\}\) to stabilize scale.

\subsubsection{Stage specific imputation via Buckley–James boosting}

Fix a stage \(k\) and a treatment option \(a\in\mathcal A\).
Let
\begin{equation}
\mathcal I_{k}(a)
=
\{i:\eta_{i,k}=1,\ A_{i,k}=a\}
\end{equation}
denote the set of participants observed at stage \(k\) and randomized to \(a\).
We fit an arm specific regression function \(f_{k,a}(H^{-}_{i,k})\) using only
\(\{(Y_{i,k},\delta_{i,k},H^{-}_{i,k}) : i\in\mathcal I_{k}(a)\}\).
Given a fitted function \(\widehat f_{k,a}\), define residuals for \(i\in\mathcal I_k(a)\) as
\begin{equation}
r_{i,k,a}
=
Y_{i,k}-\widehat f_{k,a}(H^{-}_{i,k}).
\end{equation}
Let \(\widehat F_{k,a}\) be the Kaplan Meier estimator of the residual distribution based on
\(\{(r_{i,k,a},\delta_{i,k}) : i\in\mathcal I_k(a)\}\).
For a censored observation, the Buckley–James conditional mean of the residual given \(r\) is estimated by
\begin{equation}
\widehat \mu_{k,a}(r)
=
\frac{\int_{r}^{\infty} t\, d\widehat F_{k,a}(t)}{1-\widehat F_{k,a}(r)}.
\end{equation}
The arm specific imputed stage outcome under treatment \(a\) is
\begin{equation}
\label{eq:BJ_impute}
Y^{*}_{i,k}(a)
=
\delta_{i,k}\,Y_{i,k}
+
(1-\delta_{i,k})
\Bigl\{
\widehat f_{k,a}(H^{-}_{i,k})
+
\widehat \mu_{k,a}\!\bigl(Y_{i,k}-\widehat f_{k,a}(H^{-}_{i,k})\bigr)
\Bigr\},
\end{equation}
where \(\widehat f_{k,a}\) and \(\widehat F_{k,a}\) are estimated using \(\mathcal I_k(a)\).
In counterfactual evaluation, \(\widehat f_{k,a}\) can be applied to any participant \(i\) by plugging in \(H^{-}_{i,k}\).

We consider two base learners for estimating \(\widehat f_{k,a}\) by boosting: componentwise least squares with twin boosting refinement (Algorithm~\ref{alg:bjtwin}), and regression trees (Algorithm~\ref{alg:bjtree}).

\begin{algorithm}[h!]
\caption{Buckley–James twin boosting with componentwise least squares at stage \(k\) for arm \(a\)}
\label{alg:bjtwin}
\small
\setstretch{1}
\begin{algorithmic}[1]
\State \textbf{Inputs:}
\(\{(Y_{i,k},\delta_{i,k},H^{-}_{i,k}) : i\in\mathcal I_k(a)\}\),
learning rate \(\nu\in(0,1]\),
first round boosting steps \(M_1\),
second round boosting steps \(M_2\),
maximum Buckley–James iterations \(R_{\max}\),
tolerance \(\mathrm{tol}>0\).
\State \textbf{Initialize:} set \(Y^{*(0)}_{i,k}\leftarrow Y_{i,k}\) for all \(i\in\mathcal I_k(a)\).
\State Set \(\widehat f^{(0)}_{k,a}(H)\leftarrow |\mathcal I_k(a)|^{-1}\sum_{i\in\mathcal I_k(a)} Y^{*(0)}_{i,k}\).
\For{\(R=1,2,\ldots,R_{\max}\)}
  \Statex \textbf{First round boosting.}
  \State Set \(\overline{Y}^{*(R-1)}_{k,a}\leftarrow |\mathcal I_k(a)|^{-1}\sum_{i\in\mathcal I_k(a)} Y^{*(R-1)}_{i,k}\).
  \State Initialize \(\widehat f^{(R)}_{0,k,a}(H)\leftarrow \overline{Y}^{*(R-1)}_{k,a}\) and \(\mathcal S^{(R)}_{k,a}\leftarrow\emptyset\).
  \For{\(m=0,1,\ldots,M_1-1\)}
    \State \(U_{i,m}\leftarrow Y^{*(R-1)}_{i,k}-\widehat f^{(R)}_{m,k,a}(H^{-}_{i,k})\) for all \(i\in\mathcal I_k(a)\).
    \For{\(j=1,\ldots,p\)}
      \State \(\displaystyle
      \beta_{j,m}\leftarrow
      \frac{\sum_{i\in\mathcal I_k(a)} H^{-}_{i,j,k}\,U_{i,m}}{\sum_{i\in\mathcal I_k(a)} (H^{-}_{i,j,k})^{2}}
      \),
      \(\quad
      g^{(j)}_{m}(H^{-}_{i,k})\leftarrow \beta_{j,m}H^{-}_{i,j,k}\).
      \State \(\mathrm{RSS}_{j,m}\leftarrow\sum_{i\in\mathcal I_k(a)}\bigl(U_{i,m}-g^{(j)}_{m}(H^{-}_{i,k})\bigr)^{2}\).
    \EndFor
    \State \(j^{*}\leftarrow \arg\min_{j}\mathrm{RSS}_{j,m}\), \(\quad \mathcal S^{(R)}_{k,a}\leftarrow \mathcal S^{(R)}_{k,a}\cup\{j^{*}\}\).
    \State \(\widehat f^{(R)}_{m+1,k,a}(H)\leftarrow \widehat f^{(R)}_{m,k,a}(H)+\nu\, g^{(j^{*})}_{m}(H)\).
  \EndFor
  \State Set \(\widehat f^{(R)}_{\mathrm{init},k,a}\leftarrow \widehat f^{(R)}_{M_1,k,a}\).

  \Statex \textbf{Second round twin boosting.}
  \State Initialize \(\widehat f^{(R,\mathrm{tw})}_{0,k,a}(H)\leftarrow \overline{Y}^{*(R-1)}_{k,a}\).
  \For{\(m=0,1,\ldots,M_2-1\)}
    \State \(U^{\mathrm{tw}}_{i,m}\leftarrow Y^{*(R-1)}_{i,k}-\widehat f^{(R,\mathrm{tw})}_{m,k,a}(H^{-}_{i,k})\) for all \(i\in\mathcal I_k(a)\).
    \For{\(j\in\mathcal S^{(R)}_{k,a}\)}
      \State Fit \(g^{(j)}_{m}(H^{-}_{i,k})=\beta_{j,m}H^{-}_{i,j,k}\) by least squares on \(\{U^{\mathrm{tw}}_{i,m}\}_{i\in\mathcal I_k(a)}\).
      \State \(\displaystyle
      \rho_{j,m}\leftarrow
      \widehat{\mathrm{cor}}\!\Bigl(\{g^{(j)}_{m}(H^{-}_{i,k})\}_{i\in\mathcal I_k(a)},\ \{\widehat f^{(R)}_{\mathrm{init},k,a}(H^{-}_{i,k})\}_{i\in\mathcal I_k(a)}\Bigr)^{2}
      \).
      \State \(\mathrm{PRSS}_{j,m}\leftarrow \mathrm{RSS}_{j,m}/\max(\rho_{j,m},10^{-8})\).
    \EndFor
    \State \(j^{*}\leftarrow \arg\min_{j\in\mathcal S^{(R)}_{k,a}} \mathrm{PRSS}_{j,m}\).
    \State \(\widehat f^{(R,\mathrm{tw})}_{m+1,k,a}(H)\leftarrow \widehat f^{(R,\mathrm{tw})}_{m,k,a}(H)+\nu\, g^{(j^{*})}_{m}(H)\).
  \EndFor
  \State Set \(\widehat f^{(R)}_{k,a}\leftarrow \widehat f^{(R,\mathrm{tw})}_{M_2,k,a}\).

  \Statex \textbf{Buckley–James imputation.}
  \State \(r^{(R)}_{i,k,a}\leftarrow Y_{i,k}-\widehat f^{(R)}_{k,a}(H^{-}_{i,k})\) for all \(i\in\mathcal I_k(a)\).
  \State Build Kaplan Meier \(\widehat F^{(R)}_{k,a}\) from \(\{(r^{(R)}_{i,k,a},\delta_{i,k}) : i\in\mathcal I_k(a)\}\).
  \State Update \(Y^{*(R)}_{i,k}\leftarrow \delta_{i,k}Y_{i,k}+(1-\delta_{i,k})\Bigl\{\widehat f^{(R)}_{k,a}(H^{-}_{i,k})+\frac{\int_{r^{(R)}_{i,k,a}}^{\infty} t\, d\widehat F^{(R)}_{k,a}(t)}{1-\widehat F^{(R)}_{k,a}(r^{(R)}_{i,k,a})}\Bigr\}\).

  \Statex \textbf{Convergence check.}
  \If{\(\max_{i\in\mathcal I_k(a)} \bigl|\widehat f^{(R)}_{k,a}(H^{-}_{i,k})-\widehat f^{(R-1)}_{k,a}(H^{-}_{i,k})\bigr|\le \mathrm{tol}\)}
    \State \textbf{break}
  \EndIf
\EndFor
\State \textbf{Output:} \(\widehat f_{k,a}=\widehat f^{(R)}_{k,a}\), \(\widehat F_{k,a}=\widehat F^{(R)}_{k,a}\), and imputed \(\{Y^{*(R)}_{i,k}\}_{i\in\mathcal I_k(a)}\).
\end{algorithmic}
\end{algorithm}

\begin{algorithm}[h!]
\caption{Buckley–James boosting with regression trees at stage \(k\) for arm \(a\)}
\label{alg:bjtree}
\small
\setstretch{1}
\begin{algorithmic}[1]
\State \textbf{Inputs:}
\(\{(Y_{i,k},\delta_{i,k},H^{-}_{i,k}) : i\in\mathcal I_k(a)\}\),
learning rate \(\nu\in(0,1]\),
boosting steps \(M\),
tree complexity constraints (for example maximum depth and minimum leaf size),
maximum Buckley–James iterations \(R_{\max}\),
tolerance \(\mathrm{tol}>0\).
\State \textbf{Initialize:} set \(Y^{*(0)}_{i,k}\leftarrow Y_{i,k}\) for all \(i\in\mathcal I_k(a)\).
\State Set \(\widehat f^{(0)}_{k,a}(H)\leftarrow |\mathcal I_k(a)|^{-1}\sum_{i\in\mathcal I_k(a)} Y^{*(0)}_{i,k}\).
\For{\(R=1,2,\ldots,R_{\max}\)}
  \Statex \textbf{Tree boosting fit.}
  \State Set \(\overline{Y}^{*(R-1)}_{k,a}\leftarrow |\mathcal I_k(a)|^{-1}\sum_{i\in\mathcal I_k(a)} Y^{*(R-1)}_{i,k}\).
  \State Initialize \(\widehat f^{(R)}_{0,k,a}(H)\leftarrow \overline{Y}^{*(R-1)}_{k,a}\).
  \For{\(m=0,1,\ldots,M-1\)}
    \State \(U_{i,m}\leftarrow Y^{*(R-1)}_{i,k}-\widehat f^{(R)}_{m,k,a}(H^{-}_{i,k})\) for all \(i\in\mathcal I_k(a)\).
    \State Fit regression tree \(g_m(\cdot)\) to \(\{(H^{-}_{i,k},U_{i,m}) : i\in\mathcal I_k(a)\}\) under the chosen complexity constraints.
    \State \(\widehat f^{(R)}_{m+1,k,a}(H)\leftarrow \widehat f^{(R)}_{m,k,a}(H)+\nu\, g_m(H)\).
  \EndFor
  \State Set \(\widehat f^{(R)}_{k,a}\leftarrow \widehat f^{(R)}_{M,k,a}\).

  \Statex \textbf{Buckley–James imputation.}
  \State \(r^{(R)}_{i,k,a}\leftarrow Y_{i,k}-\widehat f^{(R)}_{k,a}(H^{-}_{i,k})\) for all \(i\in\mathcal I_k(a)\).
  \State Build Kaplan Meier \(\widehat F^{(R)}_{k,a}\) from \(\{(r^{(R)}_{i,k,a},\delta_{i,k}) : i\in\mathcal I_k(a)\}\).
  \State Update \(Y^{*(R)}_{i,k}\leftarrow \delta_{i,k}Y_{i,k}+(1-\delta_{i,k})\Bigl\{\widehat f^{(R)}_{k,a}(H^{-}_{i,k})+\frac{\int_{r^{(R)}_{i,k,a}}^{\infty} t\, d\widehat F^{(R)}_{k,a}(t)}{1-\widehat F^{(R)}_{k,a}(r^{(R)}_{i,k,a})}\Bigr\}\).

  \Statex \textbf{Convergence check.}
  \If{\(\max_{i\in\mathcal I_k(a)} \bigl|\widehat f^{(R)}_{k,a}(H^{-}_{i,k})-\widehat f^{(R-1)}_{k,a}(H^{-}_{i,k})\bigr|\le \mathrm{tol}\)}
    \State \textbf{break}
  \EndIf
\EndFor
\State \textbf{Output:} \(\widehat f_{k,a}=\widehat f^{(R)}_{k,a}\), \(\widehat F_{k,a}=\widehat F^{(R)}_{k,a}\), and imputed \(\{Y^{*(R)}_{i,k}\}_{i\in\mathcal I_k(a)}\).
\end{algorithmic}
\end{algorithm}

The learning rate \(\nu\) is fixed at a small value, and we use \(\nu=0.1\) as a default.
Regularization is controlled primarily by the number of boosting steps, namely \(M\) for tree boosting and \((M_1,M_2)\) for twin boosting.
We select these iteration counts using \(V\) fold cross validation within each stage and treatment arm.
Because \(Y_{i,k}\) is censored, cross validation scores are computed using imputed responses produced by the Buckley–James iteration on the training folds, with the imputation and model fit recomputed within each fold.
For further details, see \citet{wang2010buckley}.

\subsubsection{Backward estimation of Q-functions}

We estimate Q-functions by backward recursion using the imputed stage outcomes.
At the final stage \(K\), define
\begin{equation}
\widehat Q_K\!\bigl(H_{i,K}(a_K)\bigr)
=
Y^{*}_{i,K}(a_K).
\end{equation}
For stages \(k=K-1,\ldots,1\), define the pseudo outcome under treatment \(a_k\) as
\begin{equation}
\label{eq:pseudo_outcome}
\widetilde Y_{i,k}(a_k)
=
Y^{*}_{i,k}(a_k)
+
\mathbb I(\eta_{i,k+1}=1)\,
\max_{a_{k+1}\in\mathcal A}
\widehat Q_{k+1}\!\bigl(H_{i,k+1}(a_{k+1})\bigr),
\end{equation}
so that continuation value contributes only for participants who are observed to enter stage \(k+1\).
We estimate \(\widehat Q_k\!\bigl(H_{i,k}(a_k)\bigr)\) by regressing \(\widetilde Y_{i,k}(A_{i,k})\) on \((H^{-}_{i,k},A_{i,k})\) using the same boosting learners as above.
The estimated optimal decision rule at stage \(k\) is
\begin{equation}
\widehat d_k(H^{-}_{i,k})
=
\arg\max_{a_k\in\mathcal A}\widehat Q_k\!\bigl(H_{i,k}(a_k)\bigr).
\end{equation}

\subsubsection{Identifiability assumptions}

The identifiability conditions follow standard arguments for sequentially randomized trials.
For each participant \(i\) and stage \(k\), let \(T_{i,k}(\bar a_k)\) denote the stage specific potential time under history \(\bar a_k\).
We assume:

\begin{enumerate}
\item \textbf{Consistency.}
For all \(k\) and all \(\bar a_k \in \mathcal A^k\),
\begin{equation}
\bar A_{i,k} = \bar a_k
\ \Longrightarrow\
T_{i,k}
=
T_{i,k}(\bar a_k).
\end{equation}

\item \textbf{Sequential randomization.}
For all \(k\) and all \(a_k\in\mathcal A\),
\begin{equation}
A_{i,k}\ \perp\ T_{i,k}(\bar A_{i,k-1},a_k)
\ \big|\ H^{-}_{i,k},\ \eta_{i,k}=1,
\end{equation}
where \(\bar A_{i,k-1}=(A_{i,1},\ldots,A_{i,k-1})\) denotes the realized treatment history through stage \(k-1\).

\item \textbf{Positivity.}
For all \(k\) and all \(a_k\in\mathcal A\),
\begin{equation}
\mathbb P(A_{i,k}=a_k\mid H^{-}_{i,k},\eta_{i,k}=1)>0
\quad \text{almost surely}.
\end{equation}

\item \textbf{No interference.}
For all \(i\neq j\), the potential outcomes of participant \(i\) do not depend on treatments assigned to participant \(j\).

\end{enumerate}

\section{Simulation Study}
This simulation study evaluates the performance of five Q-value estimation approaches for right-censored survival data in the context of a multistage clinical trial designed to inform dynamic treatment regimes (DTRs). 
Specifically, we compare the accuracy of treatment decisions derived from the true (oracle) Q-values, Buckley--James Q-learning with linear regression (BJ-Q), as well as its extensions using twin boosting (BJ-LS Q), regression trees (BJ-Tree Q), and Cox proportional hazards models (Cox-Q). 
The BJ-Q method follows the framework proposed by \citet{lee2025qlearninglinear}.

We simulated data for \( n \in \{500, 1000\} \) patients enrolled in a two-stage longitudinal randomized clinical trial. 
Each patient was followed over \( K = 2 \) clinical decision stages, indexed by \( k = 1, 2 \), and characterized by four covariates: sex, CD4 count, body mass index (BMI), and age. 
The binary sex variable was generated as \( \text{Sex}_i \sim \text{Bernoulli}(0.5) \). 
The CD4 count at each stage was generated from a scaled uniform distribution, 
\( \text{CD4}_{i,k} \sim \text{Unif}(1, 3) \), 
representing a normalized version of the true CD4$^+$ T-lymphocyte counts typically observed in the ACTG175 HIV clinical trial (ranging approximately from 50 to 1000 cells/µL). 
The variable was then transformed to induce nonlinearity:
\[
\text{CD4}_{i,k}^{\text{trans}} 
= \left(\text{CD4}_{i,k}\right)^{2.3} 
- \text{median}\left\{ \left(\text{CD4}_{j,k}\right)^{2.3} \right\}_{j=1}^n.
\]
Here, the CD4 count reflects immune function, serving as a key biomarker of HIV disease progression and a primary endpoint in ACTG175 for evaluating antiretroviral treatment efficacy. 
BMI and age were both generated as time-invariant covariates: 
\( \text{BMI}_i \sim \mathcal{N}(25,\, 5^2) \) and \( \text{Age}_i \sim \mathcal{N}(50,\, 10^2) \). 
At each stage \( k \), treatment \( A_{i,k} \in \{0, 1\} \) was randomly assigned with equal probability. 
The potential survival time under treatment \( a_k \) was generated from the nonlinear model:
\begin{align*}
T_{i,k}(a_k) =\ & \beta_0 + \beta_1 \cdot \text{Sex}_i + \beta_2 \cdot \text{CD4}_{i,k}^{\text{trans}} + \beta_3 \cdot \log(\text{BMI}_i) \\
& + \beta_4 \cdot \sqrt{\text{Age}_i} + \beta_5 \cdot \mathbbm{1}(a_k = 1) 
+ \beta_6 \cdot \text{CD4}_{i,k}^{\text{trans}} \cdot \mathbbm{1}(a_k = 1) + \varepsilon_{i,k},
\end{align*}
where \( \varepsilon_{i,k} \sim \mathcal{N}(0, 1) \). 
The parameters were set to 
\( \beta_0 = 10, \beta_1 = 0.4, \beta_2 = -1, \beta_3 = -0.4, \beta_4 = -0.01, \beta_5 = 0.05, \beta_6 = 1.3 \). 
Right censoring was introduced by sampling a patient-specific censoring time 
\( C_i \sim \text{Unif}(q_{0.2}(T_{i,1}),\, q_{0.8}(T_{i,1})) \) 
in the single-stage setting, and 
\( C_i \sim \text{Unif}(q_{0.2}(T_{i,1} + T_{i,2}),\, q_{0.8}(T_{i,1} + T_{i,2})) \) 
in the two-stage setting, where \( T_{i,1} \) and \( T_{i,2} \) denote the uncensored survival times at each stage. 
Here, \( q_p(\cdot) \) denotes the \( p \)-th empirical quantile computed across the simulated sample, 
to mimic random loss to follow-up and administrative censoring typically observed in clinical trials.

\subsection{Single Stage Setting}

We first consider the single-stage setting (\(k = 1\)).  
The true Q-values under each treatment arm are defined as:
\[
\begin{aligned}
Q_{i,1}^{(1)} =\ & \beta_0 + \beta_1 \cdot \text{Sex}_i + \beta_2 \cdot \text{CD4}_{i,1}^{\text{trans}} + \beta_3 \cdot \log(\text{BMI}_i) \\
& + \beta_4 \cdot \sqrt{\text{Age}_i} + \beta_5 + \beta_6 \cdot \text{CD4}_{i,1}^{\text{trans}}, \\
Q_{i,1}^{(0)} =\ & \beta_0 + \beta_1 \cdot \text{Sex}_i + \beta_2 \cdot \text{CD4}_{i,1}^{\text{trans}} + \beta_3 \cdot \log(\text{BMI}_i) \\
& + \beta_4 \cdot \sqrt{\text{Age}_i}.
\end{aligned}
\]
However, under right censoring, the observed survival times are incomplete, and special techniques are required to consistently estimate Q-functions. In the Buckley–James Boost Q-learning framework, censored outcomes are handled through iterative imputation and model fitting. The Q-functions are estimated as:
\[
\begin{aligned}
\widehat{Q}_{i,1}^{(1)} &= \hat{f}_1(\text{Sex}_i, \text{CD4}_{i,1}, \text{BMI}_i, \text{Age}_i), \\
\widehat{Q}_{i,1}^{(0)} &= \hat{f}_0(\text{Sex}_i, \text{CD4}_{i,1}, \text{BMI}_i, \text{Age}_i),
\end{aligned}
\]
where \( \hat{f}_1(\cdot) \) and \( \hat{f}_0(\cdot) \) are flexible functions fitted separately for each treatment group. 
These functions can be estimated using the classical Buckley-James linear model (BJ) or more flexible boosting-based methods, such as componentwise least squares (BJ-LS) or regression trees (BJ-Tree), both of which impute censored survival times before model fitting (see Algorithms~\ref{alg:bjtwin} and~\ref{alg:bjtree}).
For comparison, we also implemented a Cox regression--based imputation. 
At each stage \( k \), we fit a Cox proportional hazards model:
\[
\lambda_k\!\left(t \mid \text{Sex}_i, \text{CD4}_{i,k}, \text{BMI}_i, \text{Age}_i \right)
= \lambda_{0,k}(t) 
\exp\!\left( 
\beta_k^\top 
(\text{Sex}_i, \text{CD4}_{i,k}, \text{BMI}_i, \text{Age}_i)
\right),
\]
where \( \lambda_{0,k}(t) \) is the baseline hazard function and \( \beta_k \) the coefficient vector. 
The expected survival time was approximated by numerical integration of the estimated survival function:
\begin{equation}
\label{eq:cox-imputation}
\hat{Y}_{i,k}^{\text{Cox}} 
= \int_0^\infty 
\exp\!\left[
-\hat{\Lambda}_{0,k}(t)
\exp\!\left\{
\beta_k^\top 
(\text{Sex}_i, \text{CD4}_{i,k}, \text{BMI}_i, \text{Age}_i)
\right\}
\right] dt,
\end{equation}
with \( \hat{\Lambda}_{0,k}(t) \) estimated via the Breslow estimator. 
Unlike the BJ-based methods, the Cox approach relies on the proportional hazards assumption and provides a semi-parametric alternative for handling censored survival outcomes in the Q-learning framework.

\begin{figure}[h!]
  \centering
  \begin{subfigure}[b]{0.8\linewidth}
    \centering
    \includegraphics[width=1\textwidth]{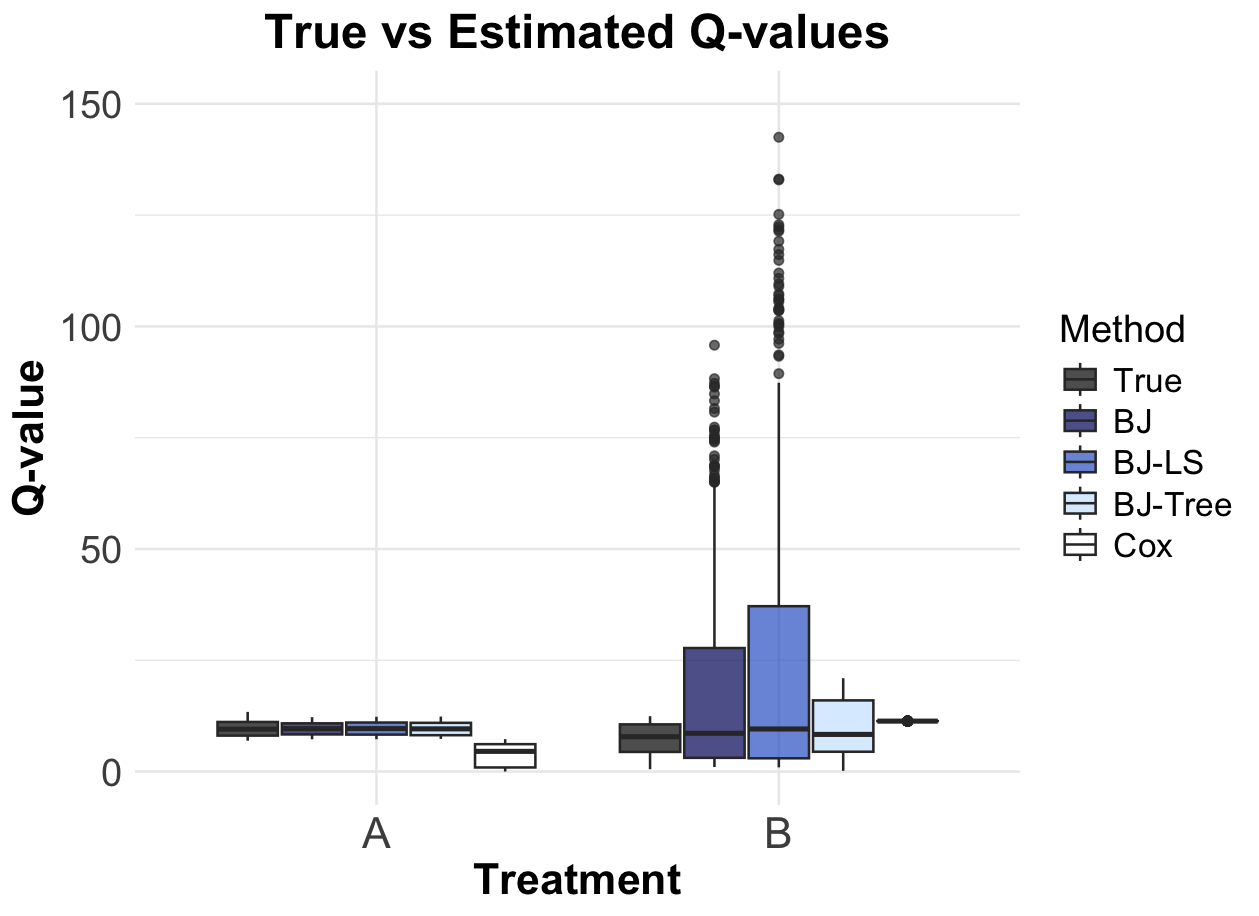}
    \caption{($n$, Censoring Rate) = (500, 0.5)}
  \end{subfigure}
  \vskip\baselineskip
  \begin{subfigure}[b]{0.8\linewidth}
    \centering
    \includegraphics[width=1\textwidth]{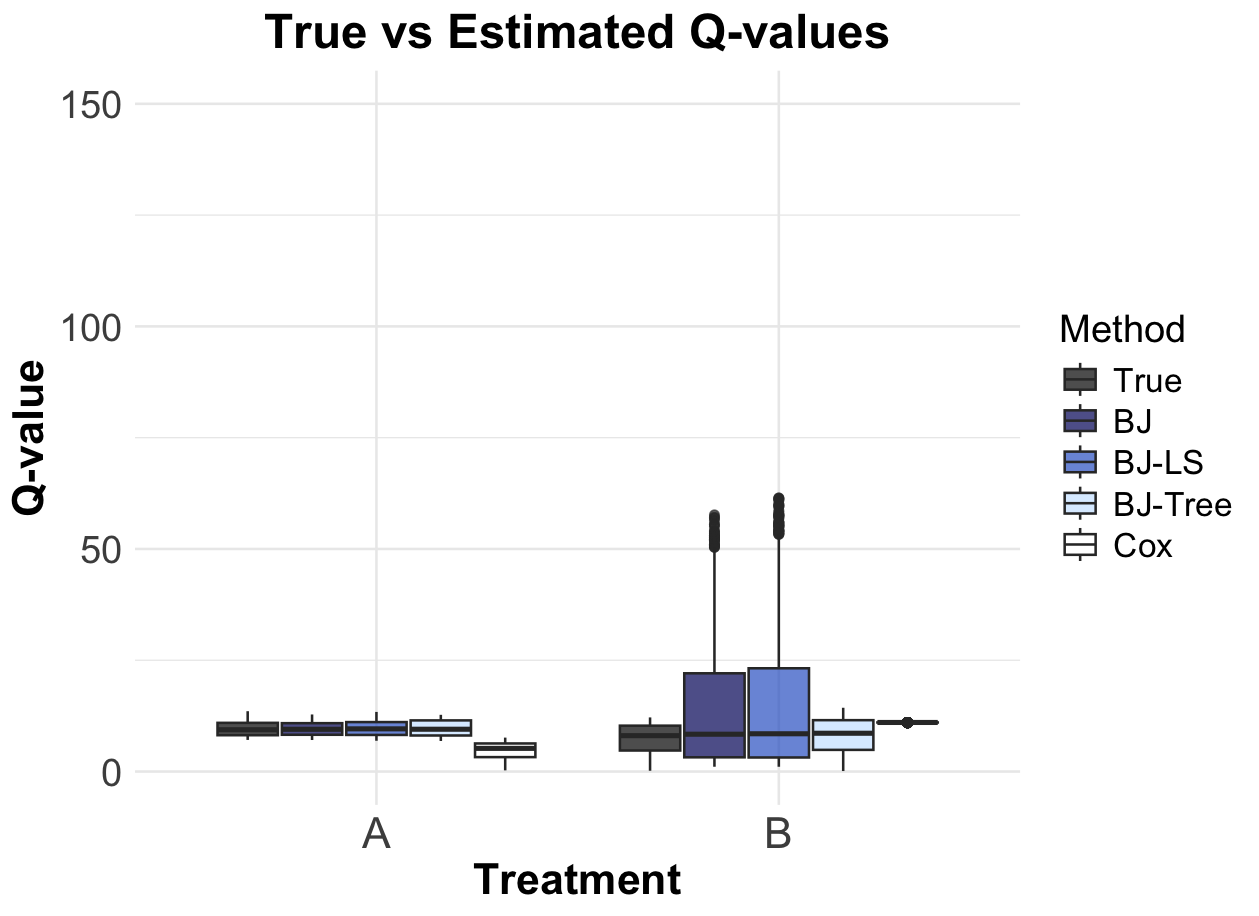}
    \caption{($n$, Censoring Rate) = (1000, 0.5)}
  \end{subfigure}
\caption{Comparison of Estimated Q-values for Treatments A and B under a Single-Stage Dynamic Treatment Regime. Each panel represents a single simulated replicate with approximately 50\% right-censoring. Boxplots display the distribution of true Q-values versus those estimated using Buckley–James Q-learning with linear regression (BJ), twin boosting (BJ-LS), regression trees (BJ-Tree), and Cox-based Q-learning (Cox), across varying sample sizes.}
  \label{fig:qvalue_comparison_single}
\end{figure}

Figure~\ref{fig:qvalue_comparison_single} presents boxplots comparing the true and estimated Q-values for Treatments~A and~B under a single-stage dynamic treatment regime, with sample sizes \( n = 500 \) and \( n = 1000 \), each subject to a 50\% censoring rate. 
The methods evaluated include the oracle (True), Buckley-James Q-learning with linear regression (BJ), twin boosting with componentwise least squares (BJ-LS), regression tree (BJ-Tree), and Cox-based Q-learning (Cox). 
Among these, BJ-Tree most closely recovers the distribution of the true Q-values, while the other methods exhibit noticeable deviations due to their limited capacity to capture nonlinear effects.

To evaluate treatment decision accuracy, we compared estimated optimal treatment assignments against oracle decisions derived from the true Q-functions prior to censoring. The true optimal decision rule assigns Treatment~A if the expected counterfactual survival under Treatment~A exceeds that under Treatment~B:
\[
d_{i,1}^{\text{true}} = \mathbbm{1}\left\{ Q_{i,1}^{(1)} > Q_{i,1}^{(0)} \right\},
\]
where \( Q_{i,1}^{(1)} \) and \( Q_{i,1}^{(0)} \) denote the true potential outcomes defined according to the known data-generating mechanism.
The estimated treatment decision rule is defined analogously based on the model-based Q-function estimates:
\[
\widehat{d}_{i,1}^{\dagger} = \mathbbm{1}\left\{ \widehat{Q}_{i,1}^{(1), \dagger} > \widehat{Q}_{i,1}^{(0), \dagger} \right\},
\]
where \( \dagger \in \{ \text{BJ}, \text{BJ-LS}, \text{BJ-Tree}, \text{Cox} \} \) denotes the estimation method used. Each method fits a separate function for each treatment group:
\[
\begin{aligned}
\widehat{Q}_{i,1}^{(1), \dagger} &= \hat{f}_1^{\dagger}(\text{Sex}_i, \text{CD4}_{i,1}, \text{BMI}_i, \text{Age}_i), \\
\widehat{Q}_{i,1}^{(0), \dagger} &= \hat{f}_0^{\dagger}(\text{Sex}_i, \text{CD4}_{i,1}, \text{BMI}_i, \text{Age}_i),
\end{aligned}
\]
where \( \hat{f}_1^{\dagger} \) and \( \hat{f}_0^{\dagger} \) are fitted Q-functions where \( \dagger \in \{ \text{BJ}, \text{BJ-LS}, \text{BJ-Tree}, \text{Cox} \} \) denotes the estimation method used.
Each rule assigns the treatment with the higher estimated survival benefit. Decision accuracy was defined as the proportion of individuals whose estimated optimal treatment matched the oracle treatment:
\[
\text{Accuracy}^{\dagger} = \frac{1}{n} \sum_{i=1}^{n} \mathbbm{1} \left\{ \widehat{d}_{i,1}^{\dagger} = d_{i,1}^{\text{true}} \right\}.
\]

\begin{table}[H]
\small
\centering
\caption{Treatment Decision Accuracy by Method and Sample Size under Single Stage Setting across 100 replications}
\begin{tabular}{llcccccc}
\toprule
\textbf{Method} & \textbf{Sample Size} & \textbf{Min} & \textbf{1st Qu.} & \textbf{Median} & \textbf{Mean} & \textbf{3rd Qu.} & \textbf{Max} \\
\midrule
BJ       & 500  & 0.8700 & 0.8720 & 0.8720 & 0.8917 & 0.9180 & 0.9240 \\
         & 1000 & 0.8840 & 0.8908 & 0.9005 & 0.9002 & 0.9100 & 0.9160 \\
BJ-LS    & 500  & 0.8540 & 0.8620 & 0.8700 & 0.8832 & 0.9120 & 0.9140 \\
         & 1000 & 0.8730 & 0.8775 & 0.8895 & 0.8882 & 0.9002 & 0.9010 \\
BJ-Tree  & 500  & 0.9000 & 0.9100 & 0.9180 & 0.9221 & 0.9400 & 0.9480 \\
         & 1000 & 0.9130 & 0.9167 & 0.9200 & 0.9285 & 0.9317 & 0.9610 \\
Cox      & 500  & 0.4240 & 0.4360 & 0.4380 & 0.4402 & 0.4500 & 0.4500 \\
         & 1000 & 0.4340 & 0.4392 & 0.4420 & 0.4405 & 0.4432 & 0.4440 \\
\bottomrule
\end{tabular}
\label{tab:accuracy-comparison-singlestage}
\end{table}

Table~\ref{tab:accuracy-comparison-singlestage} presents treatment decision accuracy across 100 replicates under a single-stage dynamic treatment regime (\(k = 1\)) for two sample sizes, \(n = 500\) and \(n = 1000\), using the Buckley–James Q-learning framework. 
Among all methods, BJ-Tree achieved the highest decision accuracy. Median accuracy exceeded 91\% for both sample sizes and improved with larger sample size, with reduced variability across replicates. This demonstrates the method’s robustness in capturing complex, nonlinear covariate-treatment interactions under right-censored survival outcomes.
BJ-LS, which implements boosting with componentwise least squares, also performed well, achieving median accuracy close to 89\% at \(n = 1000\), although consistently lower than that of BJ-Tree. 
In contrast, Cox-based Q-learning resulted in markedly lower accuracy, with median values below 45\% across both sample sizes. This poor performance likely stems from the model’s reliance on the proportional hazards assumption, which fails to capture nonlinear and heterogeneous covariate effects.

\subsection{Two-Stage Setting}

We then extended the analysis to the two-stage setting (\(k \in \{1,2\}\)), where stage-specific covariates and treatment decisions are observed at each decision point. 
We assume that all patients continue to the second stage. 
The true cumulative Q-value for a treatment sequence \((a_1, a_2)\), based on the known data-generating mechanism, is defined as:
\[
Q_i^{(a_1, a_2)} = Q_{i,1}^{(a_1)} + Q_{i,2}^{(a_2)},
\]
where \( Q_{i,k}^{(a_k)} \) denotes the counterfactual survival outcome at stage \(k\) under treatment \(a_k \in \{0,1\}\).
Correspondingly, at each stage \(k\), Q-functions were estimated separately for each treatment arm, producing model-based estimates \(\widehat{Q}_{i,k}^{(a_k), \dagger}\), where \( \dagger \) denotes the modeling approach used (e.g., BJ, BJ-LS, or BJ-Tree). 
The same method was applied at both stages to ensure consistency in estimation. The estimated cumulative Q-value under treatment sequence \((a_1, a_2)\) is then given by:
\[
\widehat{Q}_i^{(a_1, a_2), \dagger} = \widehat{Q}_{i,1}^{(a_1), \dagger} + \widehat{Q}_{i,2}^{(a_2), \dagger}.
\]

\begin{figure}[h!]
  \centering
  \begin{subfigure}[b]{0.8\linewidth}
    \centering
    \includegraphics[width=1\textwidth]{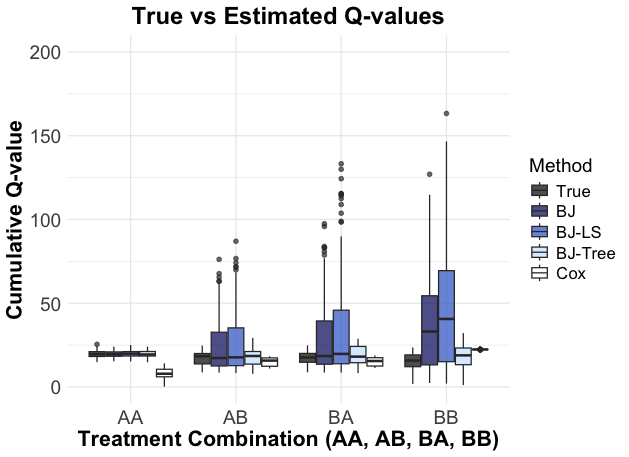}
    \caption{($n$, Censoring Rate) = (500, 0.5)}
  \end{subfigure}
  \vskip\baselineskip
  \begin{subfigure}[b]{0.8\linewidth}
    \centering
    \includegraphics[width=1\textwidth]{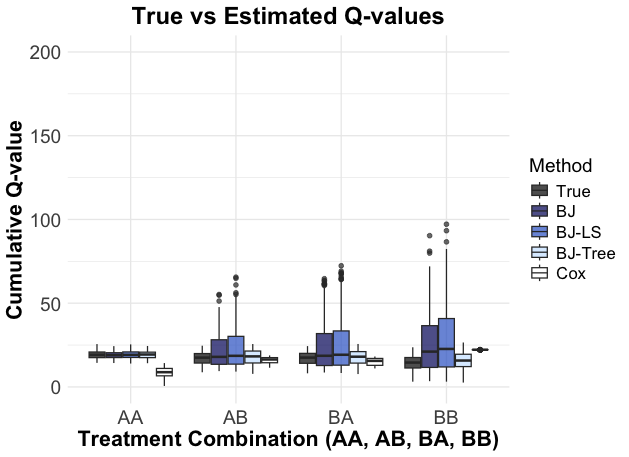}
    \caption{($n$, Censoring Rate) = (1000, 0.5)}
  \end{subfigure}
\caption{Comparison of Estimated Cumulative Q-values for Treatment Sequences $(A_1, A_2)$ under a Two-Stage Dynamic Treatment Regime. Each panel represents a single simulated replicate with approximately 50\% right-censoring at each stage. Boxplots display the distribution of true cumulative Q-values versus those estimated using Buckley–James Q-learning with linear regression (BJ), twin boosting (BJ-LS), regression trees (BJ-Tree), and Cox-based Q-learning (Cox), for each of the four treatment sequences: AA, AB, BA, and BB.}
\label{fig:qvalue_comparison_double}
\end{figure}

Figure~\ref{fig:qvalue_comparison_double} displays the distribution of estimated cumulative Q-values for the four possible treatment sequences—AA, AB, BA, and BB—under a two-stage dynamic treatment regime. The figure compares the true cumulative Q-values, derived from the known data-generating mechanism, with estimates obtained using Buckley–James Q-learning methods with linear regression (BJ), twin boosting (BJ-LS), regression trees (BJ-Tree), and Cox-based Q-learning (Cox). Among all methods, the BJ-Tree approach demonstrated the closest alignment with the true Q-value distributions across all treatment sequences, effectively capturing the nonlinear structure of the underlying survival outcomes. In contrast, the Cox-based Q-learning method consistently deviated from the truth, displaying systematic bias that reflects its limitations in modeling complex, non-proportional hazard structures inherent in the data. Notably, the magnitude of deviation from the true Q-values is greater in the two-stage setting compared to the single-stage results shown in Figure~\ref{fig:qvalue_comparison_single}, reflecting increased modeling difficulty due to compounding estimation errors across stages.

Finally, the optimal sequence is estimated by selecting the treatment pair that maximizes this total, and decision accuracy is defined similarly:
\[
\text{Accuracy} = \frac{1}{n} \sum_{i=1}^n \mathbbm{1} \left( \arg\max_{(a_1,a_2)} \widehat{Q}_i^{(a_1,a_2), \dagger} = \arg\max_{(a_1,a_2)} Q_i^{(a_1,a_2)} \right).
\]
Table~\ref{tab:accuracy-comparison-stage2} summarizes treatment decision accuracy under a two-stage dynamic treatment regime across 100 simulation replications, for sample sizes \( n = 500 \) and \( n = 1000 \). Among all methods, BJ-Tree consistently achieved the highest accuracy, with median values exceeding 84\% across both sample sizes. Accuracy further improved as the sample size increased. Both BJ and BJ-LS also performed competitively, with median accuracies above 78\%, though consistently below that of BJ-Tree. In contrast, Cox-based Q-learning demonstrated poor performance, with median accuracies below 20\% regardless of sample size, highlighting its inadequacy in settings involving nonlinear covariate-treatment interactions. These findings emphasize the importance of using flexible, nonparametric models such as BJ-Tree when estimating optimal treatment sequences under complex censoring mechanisms.

\begin{table}[H]
\small
\centering
\caption{Treatment Decision Accuracy by Method and Sample Size under Stage 2 Setting across 100 replications}
\begin{tabular}{llcccccc}
\toprule
\textbf{Method} & \textbf{Sample Size} & \textbf{Min} & \textbf{1st Qu.} & \textbf{Median} & \textbf{Mean} & \textbf{3rd Qu.} & \textbf{Max} \\
\midrule
BJ       & 500  & 0.7530 & 0.7668 & 0.7850 & 0.7876 & 0.8075 & 0.8280 \\
         & 1000 & 0.7640 & 0.7825 & 0.8020 & 0.8052 & 0.8325 & 0.8460 \\
BJ-LS    & 500  & 0.7430 & 0.7558 & 0.7780 & 0.7759 & 0.7965 & 0.8060 \\
         & 1000 & 0.7560 & 0.7620 & 0.7860 & 0.7874 & 0.8055 & 0.8260 \\
BJ-Tree  & 500  & 0.7990 & 0.8360 & 0.8440 & 0.8475 & 0.8675 & 0.8760 \\
         & 1000 & 0.7980 & 0.8380 & 0.8410 & 0.8482 & 0.8660 & 0.9000 \\
Cox      & 500  & 0.1750 & 0.1780 & 0.1875 & 0.1897 & 0.1933 & 0.2220 \\
         & 1000 & 0.1740 & 0.1920 & 0.1930 & 0.1932 & 0.2000 & 0.2020 \\
\bottomrule
\end{tabular}
\label{tab:accuracy-comparison-stage2}
\end{table}

These findings highlight the advantage of Buckley–James (BJ) boosting methods in accurately estimating optimal dynamic treatment regimes under right censoring. 
By combining flexible and iterative function estimation with modeling of censored survival outcomes, BJ boosting effectively captures individualized Q-functions without relying on restrictive hazard-based assumptions. 
Unlike the commonly used Cox-Q model, which indirectly models survival through proportional hazards and is prone to bias under model misspecification, BJ boosting directly targets conditional survival time. 
This leads to improved robustness and estimation precision. 
The benefit of BJ boosting becomes increasingly important as the number of decision stages \( K \) increases, since cumulative bias from inaccurate survival estimation can substantially reduce the accuracy of treatment decisions. 
In complex longitudinal clinical settings with right-censored data, BJ boosting methods such as twin boosting and regression tree boosting provide a powerful and interpretable framework for learning personalized and stage-specific treatment strategies.

\section{Application to the ACTG175}

We apply our proposed methodology to the analysis of the ACTG175 dataset, available from the \texttt{speff2trial} R package \citep{juraska2022package}. 
The study enrolled 2,139 HIV infected individuals randomized to one of four regimens, namely AZT monotherapy, combination therapy with AZT and didanosine (ddI), combination therapy with AZT and zalcitabine (ddC), or ddI monotherapy. 
The primary outcome \texttt{days} records time to a clinically meaningful event, including a decline in CD4 T cell count of at least 50 cells/mm\(^3\), progression to AIDS, or death. 
The censoring indicator \texttt{cens} equals 1 for observed events and 0 for censored observations. 
The dataset has a high censoring rate of approximately 75\% and includes missing values in several baseline covariates, which increases analytical complexity. 
Additional details on variables and data structure are provided in \citet{juraska2022package}. 
The primary trial objective is to compare the efficacy of monotherapy versus combination therapy among participants with baseline CD4 T cell counts between 200 and 500 cells/mm\(^3\) \citep{hammer1996trial}.

Previous analyses \citep{hammer1996trial} suggest that individuals with prior AZT exposure tend to have improved outcomes when switched to ddI, either alone or in combination with AZT, relative to continuing AZT monotherapy. 
Motivated by this clinical context, we restrict attention to the comparison between ddI monotherapy (\(A_i=0\)) and the AZT plus ddI combination regimen (\(A_i=1\)). 
We fit arm specific Q-functions under an accelerated failure time formulation using the Buckley–James boosting approach, which accommodates right censoring through iterative imputation updates and does not rely on proportional hazards assumptions. 
The estimated Q-functions are used to construct individualized treatment rules that aim to maximize each participant's predicted counterfactual survival time.

\begin{table}[!ht]
\small
\centering
\caption{Recommended treatment assignments under the Buckley--James Q-learning framework for the ACTG175 study. 
Both the twin boosting (BJ-LS) and tree boosting (BJ-Tree) methods predominantly recommend the combination therapy (\(A_i=1\)).}
\label{tab:actg_recommend}
\begin{tabular}{lcc}
\toprule
\textbf{Method} & \textbf{Recommend Treat (\(A_i=1\))} & \textbf{Recommend Control (\(A_i=0\))} \\
\midrule
BJ-LS   & 2051 & 88 \\
BJ-Tree & 2110 & 29 \\
\bottomrule
\end{tabular}
\end{table}

The individualized treatment rules derived from the Buckley--James Q-learning models strongly favored the combination regimen (\(A_i=1\)).
As shown in Table~\ref{tab:actg_recommend}, BJ-LS and BJ-Tree recommended treatment for 2051 and 2110 participants, respectively, indicating substantial agreement between the two learners.
To quantify the within-participant contrast between the two counterfactual regimes implied by the fitted Q-functions, we compared the predicted mean survival under combination therapy, \(\widehat Q_1\), with that under ddI monotherapy, \(\widehat Q_0\), for each participant.
Because \(\widehat Q_1\) and \(\widehat Q_0\) are paired counterfactual predictions evaluated on the same baseline covariates, we used the Wilcoxon signed-rank test to assess whether, across participants, the typical predicted survival under combination therapy differs from that under ddI monotherapy.
For both boosting methods, the test strongly rejected the null hypothesis that there is no systematic difference between \(\widehat Q_1\) and \(\widehat Q_0\).
Specifically, the signed-rank statistic was \(V=2{,}271{,}611\) for BJ-LS and \(V=2{,}287{,}455\) for BJ-Tree, with \(p<2.2\times 10^{-16}\) in both cases.
These results support the conclusion that the fitted models predict larger counterfactual survival under the combination regimen for the vast majority of participants, consistent with prior ACTG175 evidence reported by \citet{hammer1996trial}.

\begin{table}[!ht]
\small
\centering
\caption{Baseline covariates that are significantly associated (\(p<0.05\)) with the individualized treatment recommendations produced by the Buckley--James Q-learning framework. 
Panel~(a) presents results from the twin boosting model (BJ-LS), whereas Panel~(b) reports results from the tree based boosting model (BJ-Tree). 
Each entry shows the mean (and standard deviation) of baseline variables for participants recommended to receive the control regimen (\(A_i=0\)) versus the combination therapy (\(A_i=1\)). 
Smaller \(p\) values indicate stronger differences between these recommendation groups, highlighting baseline characteristics most strongly associated with the learned decisions.}
\label{tab:actg_split}
\begin{tabular}{lcc}
\toprule
\multicolumn{3}{c}{\textbf{(a) Buckley--James Twin Boosting (BJ-LS)}}\\
\midrule
Variable & Control vs Treat (Mean \(\pm\) SD) & \(p\) \\
\midrule
Hemoglobin (hemo)            & 0.02 (0.15) vs 0.09 (0.28) & 0.034 \\
Karnofsky score (karnof)     & 86.70 (5.62) vs 95.82 (5.62) & \(<0.001\) \\
z30 indicator (z30)          & 0.01 (0.11) vs 0.57 (0.49) & \(<0.001\) \\
Prior antiretroviral use (preanti) & 65.73 (213.93) vs 392.62 (471.94) & \(<0.001\) \\
Stratum (str2)               & 0.15 (0.36) vs 0.60 (0.49) & \(<0.001\) \\
Stratum (strat)              & 1.23 (0.58) vs 2.01 (0.90) & \(<0.001\) \\
Symptomatic (symptom)        & 0.83 (0.38) vs 0.14 (0.35) & \(<0.001\) \\
CD4 baseline (cd40)          & 325.28 (120.72) vs 351.58 (118.39) & 0.042 \\
CD8 baseline (cd80)          & 1088.09 (542.83) vs 982.27 (476.99) & 0.043 \\
\bottomrule
\end{tabular}

\vspace{1em}

\begin{tabular}{lcc}
\toprule
\multicolumn{3}{c}{\textbf{(b) Buckley--James Tree Boosting (BJ-Tree)}}\\
\midrule
Variable & Control vs Treat (Mean \(\pm\) SD) & \(p\) \\
\midrule
Age (age)                    & 39.03 (10.93) vs 35.20 (8.67) & 0.018 \\
Weight (wtkg)                & 68.13 (11.39) vs 75.22 (13.26) & 0.004 \\
Karnofsky score (karnof)     & 86.90 (7.61) vs 95.56 (5.79) & \(<0.001\) \\
Symptomatic (symptom)        & 0.93 (0.26) vs 0.16 (0.37) & \(<0.001\) \\
CD4 baseline (cd40)          & 210.38 (68.80) vs 352.43 (117.97) & \(<0.001\) \\
CD8 baseline (cd80)          & 602.41 (301.93) vs 991.91 (480.10) & \(<0.001\) \\
\bottomrule
\end{tabular}
\end{table}

Table~\ref{tab:actg_split} summarizes baseline covariates that were significantly associated with the individualized treatment recommendations derived from the Buckley--James Q-learning framework. 
Under the twin boosting model (BJ-LS), participants recommended for combination therapy had higher hemoglobin and Karnofsky scores, markedly lower symptom prevalence, and greater prior antiretroviral exposure than those recommended for ddI monotherapy. 
They also had slightly higher CD4 and lower CD8 counts at baseline, suggesting that the learned rule tended to favor combination therapy among individuals with better functional status and less symptomatic disease. 
The tree based model (BJ-Tree) yielded broadly consistent patterns, additionally identifying younger and heavier participants as more likely to be recommended to combination therapy, together with strong differences in Karnofsky score, symptom status, and baseline immune markers. 
Overall, both boosting approaches produced clinically interpretable decision rules linking disease severity, functional status, and immune function to treatment choice, and the strong preference for \(A_i=1\) is consistent with the ACTG175 evidence reported by \citet{hammer1996trial}.

\section{Application to the CALGB 8923}

We apply our proposed methodology to the CALGB 8923 dataset, available from the \texttt{mets} R package \citep{Holst2025mets}.
CALGB 8923 is a two stage randomized clinical trial conducted by the Cancer and Leukemia Group B among older patients with de novo acute myeloid leukemia \citep{stone1995granulocyte,stone2001postremission}.
In the first stage, following induction chemotherapy, participants were randomized in a double blinded manner to receive placebo or granulocyte macrophage colony stimulating factor (GM-CSF) as a study infusion \citep{stone1995granulocyte}.
We define \(A_1=0\) as placebo infusion and \(A_1=1\) as GM-CSF infusion.
Participants who achieved a complete remission (CR) were eligible for the second stage randomization, contingent on providing informed consent, with the response time recorded as \texttt{TR} \citep{stone2001postremission}.
In the second stage, these responders were rerandomized to one of two postremission regimens \citep{stone2001postremission}.
We define \(A_2=0\) as single agent cytarabine and \(A_2=1\) as cytarabine in combination with mitoxantrone (intensified postremission therapy).
For detail, see \citet{stone1995granulocyte,stone2001postremission}.

Motivated by this sequential randomization structure, we implemented a two stage Q-learning analysis under an accelerated failure time formulation.
At each stage, we fitted arm specific Q functions using the Buckley James approach, which accommodates right censoring through iterative imputation updates and does not rely on proportional hazards assumptions.
In both stages, we used the same baseline covariate set, \texttt{age}, \texttt{wbc} (white blood cell count), \texttt{sex}, and \texttt{race}.

For the second stage cohort, we restricted attention to participants who entered the second stage and had a nonmissing first stage treatment assignment.
This stage 2 cohort included 169 participants, of whom 148 deaths were observed and 21 were right censored (censoring rate 12.4\%).
The stage 2 outcome was defined as the remaining survival time after the second stage decision time, \(Y_2=\max(\texttt{time}-\texttt{TR},0.1)\), with event indicator \(\delta_2=\mathbbm{1}(\texttt{status}=1)\).
The fitted stage 2 Q functions yield paired counterfactual predictions \(\widehat Q_{2,0}\) and \(\widehat Q_{2,1}\) for each participant, corresponding to the predicted remaining survival under \(A_2=0\) and \(A_2=1\), respectively.

\begin{table}[!ht]
\small
\centering
\caption{Stage 2 recommended treatment assignments under Q-learning for the CALGB 8923 trial.}
\label{tab:calgb_stage2_recommend}
\begin{tabular}{lcc}
\toprule
\textbf{Method} & \textbf{Recommend \(A_2=1\)} & \textbf{Recommend \(A_2=0\)} \\
\midrule
Stage 2: BJ      & 95 & 74 \\
Stage 2: BJ-LS   & 96 & 73 \\
Stage 2: BJ-Tree & 95 & 74 \\
Stage 2: Cox     & 94 & 75 \\
\bottomrule
\end{tabular}
\end{table}

To evaluate whether the fitted models imply a systematic within participant difference between the two postremission options, we applied the paired Wilcoxon signed rank test to \(\widehat Q_{2,1}-\widehat Q_{2,0}\).
For BJ, the test rejected the null hypothesis that the paired differences are centered at zero (signed rank statistic \(V=4476\), \(p=2.16\times 10^{-5}\)).
For BJ-LS, the test also rejected the null hypothesis (signed rank statistic \(V=5307\), \(p=0.003\)).
For Cox, the test rejected the null hypothesis (signed rank statistic \(V=8702\), \(p=0.017\)).
Given the tendency of linear twin boosting to produce inflated counterfactual contrasts under imputation based Buckley James fitting in our simulation studies, these signals should be interpreted cautiously.
In contrast, for BJ-Tree, the paired test did not provide evidence that the paired differences depart from zero on average (\(V=6707\), \(p=0.456\)), which is consistent with the limited average survival differences reported for CALGB 8923 \citep{stone2001postremission}.
As shown in Table~\ref{tab:calgb_stage2_recommend}, all four learners produced nearly balanced recommendations at the second stage, indicating that none of the fitted models strongly favored one postremission option over the other in this cohort.
Nonetheless, the BJ-Tree recommendations varied across participants, and baseline white blood cell count differed between recommendation groups: participants for whom BJ-Tree recommended \(A_2=1\) had a lower mean baseline white blood cell count than those recommended \(A_2=0\) (mean \(20.20\) versus \(36.05\), \(p=0.022\)).

\begin{table}[!ht]
\small
\centering
\caption{Stage 1 recommended treatment assignments under Q-learning for the CALGB 8923 trial.}
\label{tab:calgb_stage1_recommend}
\begin{tabular}{lcc}
\toprule
\textbf{Method} & \textbf{Recommend \(A_1=1\)} & \textbf{Recommend \(A_1=0\)} \\
\midrule
Stage 1: BJ      & 309 & 79 \\
Stage 1: BJ-LS   & 311 & 77 \\
Stage 1: BJ-Tree & 203 & 185 \\
Stage 1: Cox     & 295 & 93 \\
\bottomrule
\end{tabular}
\end{table}

For the first stage, we included all 388 participants with an observed initial treatment assignment.
Among these participants, 181 deaths were observed and 207 were right censored (censoring rate 53.4\%).
The resulting fitted stage 1 Q functions yield paired counterfactual predictions \(\widehat Q_{1,0}\) and \(\widehat Q_{1,1}\), corresponding to the predicted overall survival under the two first stage options \(A_1=0\) and \(A_1=1\).
We again applied the paired Wilcoxon signed rank test to \(\widehat Q_{1,1}-\widehat Q_{1,0}\).
For BJ, the test strongly rejected the null hypothesis that the paired differences are centered at zero (signed rank statistic \(V=64{,}757\), \(p<2.2\times10^{-16}\)), and the resulting decision rule favored \(A_1=1\) for most participants.
For BJ-LS, the test also strongly rejected the null hypothesis (signed rank statistic \(V=62{,}966\), \(p<2.2\times10^{-16}\)), yielding a similarly decisive recommendation pattern.
For Cox, the test strongly rejected the null hypothesis (signed rank statistic \(V=58{,}717\), \(p<2.2\times10^{-16}\)), and it favored \(A_1=1\) for most participants (Table~\ref{tab:calgb_stage1_recommend}).
In contrast, for BJ-Tree, the test did not provide evidence that the paired differences depart from zero on average (\(V=39{,}314\), \(p=0.475\)), yielding a substantially more balanced recommendation pattern (Table~\ref{tab:calgb_stage1_recommend}).
Clinically, these findings suggest that the tree based learner does not identify a pronounced advantage of one induction strategy over the other in this cohort, consistent with the limited average survival differences reported for CALGB 8923 \citep{stone1995granulocyte}.
Nonetheless, the BJ-Tree recommendations varied across participants: compared with those recommended \(A_1=0\), participants recommended \(A_1=1\) were older on average (mean \(70.41\) versus \(68.59\), \(p=0.002\)) and had a lower mean baseline white blood cell count (mean \(26.40\) versus \(38.30\), \(p=0.025\)).

Overall, the CALGB 8923 analysis illustrates that the choice of learner within the Buckley James Q-learning framework can influence the implied within participant counterfactual contrasts and the resulting individualized treatment rules.
In this dataset, BJ, BJ-LS, and Cox produced statistically detectable paired differences at both stages and more decisive recommendations at the first stage, whereas BJ-Tree yielded more balanced recommendations and results that align more closely with the limited average survival differences reported in the trial \citep{stone1995granulocyte,stone2001postremission}.

\section{Discussion}

The BJ Boost Q-learning framework demonstrates strong empirical performance in estimating optimal dynamic treatment regimes under right-censored data. Our simulation study, conducted under a moderate censoring rate of approximately 50\%, showed that the method achieves high decision accuracy and reliable Q-value estimation across multiple model specifications. Among these, BJ-Tree learner consistently delivered the most accurate results, underscoring its strength in modeling nonlinear treatment effects.

Despite these promising results, limitations remain. The framework’s reliability under higher levels of censoring has not been systematically assessed. Since the Buckley–James estimator relies on imputation of censored outcomes, excessive censoring could degrade the quality of the imputed values and compromise downstream Q-learning performance. Future studies should examine the method’s robustness under various censoring scenarios.

Model misspecification also presents a concern, particularly in the early stages of the boosting process. Although regression trees offer flexibility, early-stage bias may be propagated through successive boosting iterations. Investigating regularization techniques or early stopping criteria may improve robustness in such cases.

Lastly, this study assumes randomized treatment assignment. In observational settings, where confounding is a key challenge, the BJ Boost Q-learning framework should be extended to incorporate causal adjustment techniques such as inverse probability of treatment weighting (IPTW) or propensity score matching. These enhancements would expand the method’s applicability to real-world clinical data where treatment allocation is nonrandom.

\section*{Data and Code}
The implementation of our method, along with detailed information on data generation and real data analysis, is available at \url{https://github.com/jeongjin95/BJ-Boost-Q-learning} for reproducibility.

\section*{Acknowledgments}
We thank Professor Ji-Hyun Lee from the Department of Biostatistics at the University of Florida for her valuable guidance on survival analysis and real data analysis.
We also express our sincere gratitude to the Associate Editor, the reviewers, and the Editor for their constructive comments and helpful suggestions, which greatly improved the quality of this manuscript.

\section*{Conflict of interest}
The authors declare that they have no conflict of interest.

\bibliographystyle{agsm}
\bibliography{interacttfqsample}

\end{document}